\pdfoutput=1
\PassOptionsToPackage{square,comma,numbers,sort&compress,super}{natbib}  
\documentclass{article}
\usepackage{arxiv}
\usepackage[utf8]{inputenc} 
\usepackage[T1]{fontenc}    
\usepackage{hyperref}       
\usepackage{url}            
\usepackage{booktabs}       
\usepackage{amsfonts}       
\usepackage{nicefrac}       
\usepackage{microtype}      
\usepackage{lipsum}

\usepackage{graphicx}
\usepackage{ifthen}
\usepackage{float}
\usepackage{mathtools}
\usepackage{multicol}
\usepackage{subfigure}
\usepackage{xspace}
\usepackage{multirow}
\usepackage{enumitem} 
\usepackage[algo2e, ruled, vlined, linesnumbered]{algorithm2e}
\usepackage{ulem} 
\usepackage{booktabs} 

\newcommand{\dpp}{DPP-Net\xspace}
\newcommand{\monas}{MONAS\xspace}
\newcommand{\rn}{robot network\xspace}
\newcommand{\tn}{child network\xspace}

\newcommand{\hp}{hyperparameter\xspace}
\newcommand{\hps}{hyperparameters\xspace}

\title{Searching Toward Pareto-Optimal Device-Aware Neural Architectures}

\author{
An-Chieh Cheng$^\dagger$, Jin-Dong Dong$^\dagger$, Chi-Hung Hsu$^\dagger$, Shu-Huan Chang$^\dagger$, Min Sun$^\dagger$, \\ 
\textbf{ Shih-Chieh Chang$^\dagger$, Jia-Yu Pan$^\ddagger$, Yu-Ting Chen$^\ddagger$, Wei Wei$^\ddagger$, Da-Cheng Juan$^\ddagger$ } \\
$^\dagger$ National Tsing-Hua University, Hsinchu, Taiwan \\
$^\ddagger$ Google Research, Mountain View, CA, USA \\
\texttt{\{accheng.tw, mark840205, charles1994608,tommy610240\}@gmail.com} \\
\texttt{\{sunmin@ee, scchang@cs\}.nthu.edu.tw} \\
\texttt{\{jypan, yutingchen, wewei, dacheng\}@google.com}
}

\begin{document}
\maketitle

\begin{abstract}
Recent breakthroughs in Neural Architectural Search (NAS) have achieved state-of-the-art performance in many tasks such as image classification and language understanding. However, most existing works only optimize for model accuracy and largely ignore other important factors imposed by the underlying hardware and devices, such as latency and energy, when making  inference. In this paper, we first introduce the problem of NAS and provide a survey on recent works. Then we deep dive into two recent advancements on extending NAS into multiple-objective frameworks: \monas~\cite{hsu2018monas} and \dpp~\cite{dong2018dpp}. Both \monas and \dpp are capable of optimizing accuracy and other objectives imposed by devices, searching for neural architectures that can be best deployed on a wide spectrum of devices: from embedded systems and mobile devices to workstations. Experimental results are poised to show that architectures found by \monas and \dpp achieves Pareto optimality w.r.t the given objectives for various devices.
\end{abstract}


\pdfoutput=1
\section{Introduction}

\begin{table}[t]
\caption{Comparisons of Neural Architecture Search Approaches.}
\label{tab:allapproaches}
\resizebox{1\columnwidth}{!}{%
\begin{tabular}{llllll}
\hline
\multicolumn{6}{c}{\textbf{Single-Objective Neural Architecture Search}}                                                                                                                                                                                                                                                                                                                                                              \\ \hline
\multirow{2}{*}{Approach}                                                                            & \multirow{2}{*}{\begin{tabular}[c]{@{}l@{}}Search \\ Space\end{tabular}} & \multirow{2}{*}{Algorithm} & \multirow{2}{*}{Acceleration Techniques}      & \multirow{2}{*}{\begin{tabular}[c]{@{}l@{}}Search Cost \\ (GPU Days)\end{tabular}} & \multirow{2}{*}{\begin{tabular}[c]{@{}l@{}}Additional \\ Objectives\end{tabular}} \\
                                                                                                     &                                                                          &                            &                                               &                                                                                    &                                                                                   \\ \hline
NAS\cite{zoph2016neural}                                                       & Macro        & RL        & -                                              & 22400       & -                                 \\
NasNet\cite{zoph2017learning}                                                 & Micro         & RL        & -                                             & 1800        & -                                 \\
Hierarchical\cite{liu2017hierarchical}                                        & Micro         & EA/RS     & -                                             & 300         & -                                 \\
MetaQNN\cite{baker2016designing}                                              & Macro        & RL        & -                                             & 100         & -                                 \\
GeNet\cite{xie2017genetic}                                                    & Macro        & EA        & -                                             & 17          & -                                 \\
Large-Scale\cite{real2017large}                                               & Macro        & EA        & Weight-Sharing                                & 2500        & -                                 \\
Amoeba\cite{real2018regularized}                                              & Micro         & EA        & Weight-Sharing                                & 3150        & -                                 \\
SMASH\cite{brock2017smash}                                                    & Macro        & RS        & Weight-Sharing                                & 1.5         & -                                 \\
EAS\cite{cai2018efficient}, TreeCell\cite{cai2018path} & Entire       & RL        & Weight-Sharing                                                       & 10, 200     & -                                 \\
BlockQNN\cite{zhong2017practical}                                             & Micro         & RL        & Improved Proxy , Weight-Sharing               & 96          & -                                 \\
ENAS\cite{pham2018efficient}                                                  & Macro/Micro  & RL        & Weight-Sharing                                & 1           & -                                 \\
NASH\cite{elsken2017simple}                                                   & Macro        & RS        & Weight-Sharing                                & 0.5         & -                                 \\
PNAS\cite{liu2017progressive}                                                 & Micro         & EA        & Improved Proxy                                & 150         & -                                 \\
Bender et al.~\cite{bender2018understanding}                                  & Micro         & RS        & Weight-Sharing                               & 4           & -                                 \\
DARTS\cite{liu2018darts}                                                      & Micro         & GD        & Weight-Sharing                                & 4           & -                                 \\ \hline
\multicolumn{6}{c}{\textbf{Multi-Objective Neural Architecture Search}}                                                                                                                                                                                                                                                                                                                                                               \\ \hline
DPP-Net\cite{dong2018dpp}                                                     & Micro          & EA        & Improved Proxy                                & 2                          & latency, \# params, FLOPS, memory   \\
MONAS\cite{hsu2018monas}                                                      & Macro        & RL        & -                                             & -          & power consumption                   \\
LEMONADE\cite{elsken2018multi}                                                & Macro/Micro  & EA        & Weight-Sharing                                & 56                         & \# params                           \\
Smithson et al.\cite{smithson2016neural}                                      & Macro        & EA        & Improved Proxy                                & 3                          & \# params, FLOPS                    \\
NEMO\cite{kimnemo}                                                            & Macro        & EA        & -                                             & -                          & latency                             \\
RENA~\cite{zhou2018resource}                                                  & Macro/Micro  & RL        & Weight-Sharing                               & -                          & \# params, FLOPS, compute intensity \\
MnasNet\cite{tan2018mnasnet}                                                  & Micro         & RL        & -                                             & -                          & latency
\end{tabular}
}
\end{table}

In recent years, deep neural networks (DNNs) have demonstrated impressive performance on challenging tasks such as image recognition \cite{krizhevsky2012imagenet}, speech recognition \cite{hannun2014deep}, machine translation and natural language understanding \cite{sutskever2014sequence}. Despite these great successes achieved by DNNs, designing neural architectures is usually a manual and time-consuming process that heavily relies on experience and expertise. Recently, neural architecture search (NAS) has been proposed to address this issue \cite{negrinho2017deeparchitect,zoph2016neural}. Models designed by NAS have achieved impressive performance that is close to or even outperforms the current state-of-the-art designed by domain experts in several challenging tasks \cite{zoph2017learning,pham2018efficient}, demonstrating strong promises in automating the designs of neural networks.

However, most existing works of NAS only focus on optimizing model accuracy and largely ignore other important factors (or constraints) imposed by underlying hardware and devices. For example, from workstations, mobile devices to embedded systems, each device has different computing resource and environment. Therefore, a state-of-the-art model that achieves excellent accuracy may not be suitable, or even feasible, for being deployed on certain (e.g., battery-driven) computing devices, such as mobile phones.

To this end, several works have been proposed to extend NAS into multiple objectives \cite{kimnemo,smithson2016neural,tan2018mnasnet,elsken2018multi,zhou2018resource,hsu2018monas,dong2018dpp}, as opposed to the original single objective (i.e., accuracy), to search for device-aware neural architectures. With multi-objective NAS, factors or constraints imposed by underlying physical devices can be accounted by being formulated as the corresponding objectives. Therefore, instead of finding the ``best'' model in terms of accuracy, most of these works embrace the concept of ``Pareto optimality'' w.r.t. the given objectives, which means none of the objectives can be further improved without worsening some of the other objectives.

This paper brings the following contributions:
\begin{itemize}
    \item{We survey the recent literature on NAS and summarize the proposed approaches into four categories: (a) reinforcement-learning-based, (b) evolutionary-algorithm-based, (c) search acceleration, and (d) multi-objective search.}
    \item{We deep dive into two recent advancements of multi-objective NAS---\monas and~\dpp---that effectively search for device-aware neural architectures.}
    \item{The performances of neural architectures found by~\monas and~\dpp are evaluated on a wide spectrum of devices: from a workstation, mobile devices to embedded systems, by using both accuracy and device-related metrics such as latency and energy consumption.}
\end{itemize}

The remainder of this paper is organized as follows. Section~\ref{sec:Neural Architecture Search} provides the problem definition of NAS, and surveys the related works. Section~\ref{sec:monas} deep dives into \monas, multi-objective NAS based on reinforcement learning (RL), and Section~\ref{sec:dpp} provides the details of \dpp, multi-objective NAS based on evolutionary algorithms. Finally, Section~\ref{sec:discussions} discusses about multi-objective NAS and points to possible future directions.

\section{Neural Architecture Search}
\label{sec:Neural Architecture Search}
In this section, we survey the recent literatures on NAS and summarize them into four categories: (a) reinforcement-learning based methods, (b) evolutionary-algorithm based methods, (c) search acceleration, and (d) multi-objective search. Table.~\ref{tab:allapproaches} provides the overview and comparisons among these literatures. 

\subsection{Problem Definition}
\label{sec:sec:Problem Definition}
Generally, the problem of neural architecture search can be formulated into two sub-problems: design ``Search Space'' and ``Search Algorithm''~\cite{elsken2018neural}.
\paragraph{\textbf{Search Space}}As its name suggests, search space represents a set of possible neural networks available to be searched over. Usually, a search space has a numerical representation that contains:
\begin{itemize}
    \item{Structure of a neural network, such as the depth of a neural net (i.e., the number of hidden layers) and the width of a particular hidden layer.}
    \item{Configurations, such as operation/connection types, kernel size, the number of filters.}
\end{itemize}
Therefore, given a search-space design, each neural network can be encoded into the corresponding representation. This type of search space is referred to as \textit{macro} design space. Several works of literature have been proposed \cite{saxena2016convolutional,mendoza2016towards,baker2016designing} with such macro search space of fixed network depth. Recently, Cai et al.~\cite{cai2018efficient} designed a transformable search space that enables searching over different network depths.

In addition to the conventional architectures, ``multi-branch'' structures also play an increasingly important role, especially in recent state-of-the-art convolutional neural architectures. Two prior arts, ResNet~\cite{he2016deep} and DenseNet~\cite{huang2017densely}, proposed skip-connection and dense-connection, respectively, to create ``branches'' of the data flow in a neural network. Possibly inspired by these structures, Zoph et al.~\cite{zoph2016neural} proposed to design the search space including skip connections; this search space has been quickly adopted by other works \cite{zoph2017learning,brock2017smash,real2017large,cai2018path}.

Another recent trend is to design a search space that covers only one basic \textit{cell} that will be used as the building block for constructing an entire network. This type of space designs is referred to as \textit{micro} design space, where the search cost and complexity usually can be reduced significantly. \cite{zoph2017learning} is the first work that proposed this concept. In addition to reducing the search complexity, the best-found cell can also be generalized to other tasks more easily by simply changing the number of the cells stacked \cite{real2017large,real2018regularized,liu2017progressive,pham2018efficient,cai2018path,dong2018dpp,liu2017hierarchical}. The potential drawbacks of searching for a cell structure (instead of the entire network) are in two folds: (a) the search space is usually smaller and more constrained, in which even a random search can sometimes achieve comparable results. (b) The cell structure implicates experts' design bias, which might reduce the possibility of finding truly novel and surprising architectures.

\paragraph{\textbf{Search Algorithm}}
A search algorithm is usually an iterative process that determines how a search space will be explored in order. In each step or iteration of the search process, a ``sample'' is drawn (or generated) from a search space to form a neural network, referred to as a ``child network.'' All child networks are trained on the training datasets and their accuracy on the validation datasets are then treated as the objective (or as the reward in reinforcement learning) to be optimized. The goal of a search algorithm is to find the best child network that optimizes the objective, such as minimizing the validation loss or maximizing rewards. We provide detailed explanations and discussions for different types of search algorithms in the following sections. 

\subsection{Reinforcement-Learning-Based Approaches}
\label{sec:sec:Reinforcement Learning Based Approach}
Reinforcement-learning-based approaches have been the mainstream methods for NAS, especially after Zoph et al.~\cite{zoph2016neural} demonstrated the impressive experimental results that outperform the state-of-the-art models designed by domain experts.

\paragraph{\textbf{NAS formulated as reinforcement learning (RL)}} There are three fundamental elements in RL: (a) an agent, (b) an environment, and (c) a reward. The goal is to learn the action policy for the agent to interact with the environment so that the maximum long-term rewards will be received. The interactions between the agent and the environment can be viewed as a sequential decision-making process: at each time $t$, the agent chooses an action $a_{t}$ (from the set of available actions) to interact with the environment and receives a reward. To frame NAS as an RL problem, the agent's action is to select or generate a child network, while the validation performance is taken as the reward. 

\paragraph{\textbf{Related literatures.}} In general, various RL-based approaches for NAS differ in (a) how the action space is designed, and (b) how the action policy is updated. Zoph et al.~\cite{zoph2016neural} first applied policy gradient to update the policy, and in their later work~\cite{zoph2017learning} changed to use proximal policy optimization; Baker et al.~\cite{baker2016designing} used Q-learning to update the action policy. There are also works designing action spaces differently. While most of the previous works defined actions as selecting the configuration of a new architecture, in \cite{cai2018efficient,cai2018path} the actions are defined as the operations to transform a network by adding, deleting, or widening an existing network layer.

In the sequential decision-making process, the trade-off between the exploration of new possibilities and the exploitation of old certainties determines the overall search cost. When exploring a high-dimensional search space, the search cost of RL-based approaches can be extremely high. In \cite{zoph2016neural}, the search process took 28 days with 800 GPUs to yield promising results. The search still took 4 days with 450 GPUs even it is on a simplified search space \cite{zoph2017learning}. To tackle this problem, many recent advancements on neural architecture search focus on reducing the computational cost, which will further be discussed in Section~\ref{sec:sec:Multiple Objective Architecture Search}.

\subsection{Evolutionary-Algorithm-Based Approaches}
\label{sec:sec:Evolutionary Algorithm Based Approach}

The goal of NAS, in its nature, can also be approached by the process of natural selection. Recent EA-based NAS approaches focus on searching for architectures and updating the connection weights through back-propagation.

\paragraph{\textbf{NAS formulated as evolutionary algorithm (EA)}}Evolutionary algorithms (EA) evolves a \textit{population} of models over evolution steps.
Every model in the population is a trained network and considered as an \textit{individual}.
Similar to the RL approach, the model's performance (e.g., accuracy) on the validation dataset is the measure of the quality of each individual.
At an evolution step, one or more individuals are chosen as the \textit{parent models} based on their quality. 
A copy of the parents, which is regarded as \textit{child} networks, will subsequently be created and applied with mutate operations.
After the child network is trained and evaluated on the validation set, it will be added to the population.
To sample a parent from the population, Real et al.~\cite{real2017large} adopted the tournament selection~\cite{goldberg1991comparative} which uses repeated pairwise competitions of random individuals instead of the whole population to increase the search efficiency. Most of the later works~\cite{real2018regularized,liu2017hierarchical} followed this concept and used tournament-liked selections.

\paragraph{\textbf{Related literatures.}}

One drawback of the EA algorithm is that the evolution process is usually considered to be unstable, and the quality of the final population can vary due to random mutations. Chen et al.~\cite{chen2018reinforced} proposed an RL controller to make decisions for mutation instead of doing random mutations to stabilize the search process.

To tackle this problem, Liu et al. and Dong et al.~\cite{liu2017progressive,dong2018dpp} adopted a sequential-model based optimization (SMBO) that has a similar concept to EA. Through predicting the performance of candidates, the algorithm can decide which new node to expand. We will further discuss this method in Section~\ref{sec:dpp}.

\subsection{Search Acceleration}
In both RL-based and EA-based approaches, every child network needs to be trained and evaluated in order to guide the search process. However, training each network from scratch requires significant computing resources and time (e.g., 20,000+ GPU days)~\cite{zoph2016neural}. One general speedup approach in NAS is to find proxy metrics that approximate the performance after full training (e.g., shorter training epochs~\cite{zoph2016neural}, simpler datasets~\cite{zoph2017learning}). Advanced techniques can be categorized into two types: (a) improved proxy or (b) weight-sharing. 

\paragraph{\textbf{Improved proxy.}} When using proxy metrics, the relative ranking among child networks needs to remain correlated with final accuracies in order to obtain a better result. Otherwise, a ``good'' child network may, unfortunately, have lower accuracy than a ``bad'' child network. Zhong et al.~\cite{zhong2017practical} observed that FLOPs and model size of a child network have a negative correlation with the final accuracy, and introduced a correction function applied on reward calculation with child networks' accuracies obtained by early stopping, to bridge the gap between the proxy and true accuracy. Several approaches proposed to improve proxy metrics by ``predicting'' the accuracies of neural architectures \cite{smithson2016neural,domhan2015speeding,baker2018accelerating,liu2017progressive,dong2018dpp}. Child networks predicted to have poor accuracies will be either suspended from training or directly abandoned. Domhan et al.~\cite{domhan2015speeding} proposed to make a prediction based on the learning curve of child networks. Baker et al.~\cite{baker2018accelerating} used regression models to predict the final performance of partially trained model using features based on network configurations and validation curves. Liu et al. and Dong et. at~\cite{liu2017progressive,dong2018dpp} trained surrogate models to predict the accuracies of child networks based on progressively architectural properties. The biggest challenge for predicting accuracies is the training data were scarce and costly (e.g, each sample here is the accuracy of a trained child network).

\paragraph{\textbf{Weight-sharing.}} Weight-sharing is another approach to expedites the progress of architecture search. During the neural evolution process, Real et al.~\cite{real2017large} allowed the children to inherit the parent's weights whenever possible. Inspired by \cite{real2017large}, Pham et. at~\cite{pham2018efficient} improved the efficiency of \cite{zoph2017learning} by forcing all child models to share weights instead of training each child model from scratch. The search progress is reduced to less than 16 hours with single Nvidia GTX 1080Ti GPU, which is more than 1000x GPU time reduction compared to \cite{zoph2017learning}. Brock et al.~\cite{brock2017smash} proposed one-shot model architecture search, which designs a ``main'' model with an auxiliary hypernetwork to generate the weights of the main model conditioned on the model's architecture. This results in significant speed-up for architecture search since no training for child networks is required. While this approach uses weights sampled from a distribution represented by the hypernetwork, Bender et al.~\cite{bender2018understanding} proposed to use strictly shared weights on one-shot architecture search, which trains a one-shot model which represents a wide variety of candidate architectures, then randomly evaluate these candidate architectures on the validation set using the pre-trained one-shot model weights. Similar to ~\cite{pham2018efficient,bender2018understanding}, Liu et al.~\cite{liu2018darts} trained all the weights with a one-shot model containing entire search space, and at the same time, they use gradient descent (GD) to optimize the distribution over candidate architectures. Several other works~\cite{cai2018efficient,cai2018path,jin2018efficient} explore the architecture space by network transformation/morphism, which modified a trained neural network into a new architecture using the operations such as inserting a layer or adding a skip-connection. Since network transformation/morphism begins from an existing trained network, the weights are reused and only a few more iterations of training are required to further train the new architecture.

\subsection{Multi-objective Neural Architecture Search}
\label{sec:sec:Multiple Objective Architecture Search}

Most existing methods focus only on optimizing a single objective: model accuracy. However, these models may not be suitable, or even feasible, for being deployed on certain (e.g., battery-driven) computing devices, such as mobile phones. Therefore, one recent research direction of NAS is extending NAS into multi-objective problems. In a multi-objective optimization framework, the concept of ``Pareto optimization'' is used to search for the best solution. A feasible solution is considered Pareto optimal if \textit{none} of the objectives can be improved without worsening some of the other objectives; in this situation, these solutions achieve Pareto-optimality and are referred to be on the Pareto front.

In addition to model accuracy, there are several important device-related objectives that need to be considered during NAS: inference latency, energy consumption, power consumption, memory usage, and floating-point operations (FLOPs). As Table~\ref{tab:allapproaches} shows, Elsken et al.~\cite{elsken2018multi}, Smithson et al.~\cite{smithson2016neural}, and Zhou et al.~\cite{zhou2018resource} used FLOPs and the number of parameters as the proxies of computational costs; Kim et al.~\cite{kimnemo} and Tan et al.~\cite{tan2018mnasnet} directly included actual runtime as an objective to be minimized; Hsu et al.~\cite{hsu2018monas} designed novel reward functions accounting for peak power and average energy consumption; Dong et al.~\cite{dong2018dpp} proposed to consider both device-agnostic objectives (e.g., number of parameters, FLOPs) and device-related objectives (e.g., inference latency, memory usage) using Pareto optimization.

Most of approaches for multi-objective NAS  can, again, be categorized into two type: (a) RL-based approaches~\cite{tan2018mnasnet,hsu2018monas,zhou2018resource} and (b) EA-based approaches~\cite{smithson2016neural,elsken2018multi,kimnemo,dong2018dpp}. In Section~\ref{sec:monas} and Section~\ref{sec:dpp}, we will discuss the recent advancement for each of these two types (RL and EA).

\pdfoutput=1
\section{RL-Based Multi-Objective NAS}
\label{sec:monas}

We introduce a recent advancement of RL-based approach for multi-Objective NAS: \monas~\cite{hsu2018monas}.
\monas designs different reward functions that consider both model accuracy and power consumption when exploring neural architectures. 

\subsection{Overview}
\label{sec:sec:overview}

\monas framework is built on top of a two-stage reinforcement-learning framework similar to NAS \citep{zoph2017learning}.
In the generation stage, a RNN is used as a \textit{\rn}, which generates a \hp sequence for a CNN. In the evaluation stage, \monas trains a \textit{\tn}\footnote{In \monas paper~\cite{hsu2018monas}, the authors called a \tn ``target network.''} according to the \hps outputted by the RNN. Both the accuracy and energy consumption of the \tn are used as rewards for \rn.  The \rn updates the policy generating \hp with rewards and policy gradient.

\subsection {Reward Function}
\label{sec:sec:Reward Function}

\setlength{\abovedisplayskip}{3pt}
\setlength{\belowdisplayskip}{3pt}

\monas considers three different objectives that account for the validation accuracy, the peak power, and average energy when the trained model is making inference. These objectives are then formulated as reward functions:

\begin{itemize}
\setlength\itemsep{0em}
\item \textbf{Mixed Reward:} the reward is calculated as the weighted combination of accuracy and energy; depending on the selection of $\alpha$, \rn will design a more accurate child network or a more energy-efficient network. When $\alpha=1$, \monas is degenerated to NAS proposed by Zoph et al.~\cite{zoph2016neural}. 
\begin{equation}
\label{eq:mix}
R = \alpha*Accuracy - (1-\alpha)*Energy ,\quad  \alpha \in [0, 1]
\end{equation}
\item \textbf{Power Constraint:} To search for neural networks that work under a predefined power budget, the peak power (when in serving, i.e., making inference) is formulated as a hard constraint here:
\begin{equation}
\label{eq:pow}
  R = 
  \left\{
   \begin{aligned}
   &Accuracy,&\quad &\mbox{if power} \leq \mbox{power budget}\\
   &0                                     ,&\quad &\mbox{otherwise}\\
   \end{aligned}
  \right.
\end{equation}
\item \textbf{Accuracy Constraint:} \monas also demonstrates a scenario to search for a energy-efficient model when accuracy hits (and above) the threshold:
\begin{equation}
\label{eq:accu}
  R = 
  \left\{
   \begin{aligned}
   &1 - Energy^{*}, &\quad &\mbox{if accuracy} \geq \mbox{threshold}\\
   &0                                     , &\quad &\mbox{otherwise}\\
   \end{aligned}
  \right.
\end{equation}
\end{itemize}

\noindent Note that $Energy^{*}$ in Eq(\ref{eq:accu}) means normalized energy consumption, where $Energy^{*}$ $\in [0, 1]$. These rewards will then be plugged into the RL framework described in Section~\ref{sec:sec:Reinforcement Learning Based Approach}.


\subsection{Experimental Results}
\label{sec:sec:monas-results}

All experiments in \monas are conducted on a workstation (WS) with Intel XEON E5-2620v4 processor equipped with NVIDIA GTX-1080Ti GPU cards. The NVIDIA profiling tool, nvprof, is used to measure the peak power, average power, and the runtime of CUDA kernel functions used by the \tn. The dataset used is CIFAR-10. All the experimental results in this section are excerpted from the original \monas paper (Hsu et al.~\cite{hsu2018monas}).

\begin{figure}[H]
\centering
\subfigure[Max Power 70W vs Random Search]
{\includegraphics[width=0.45\textwidth]{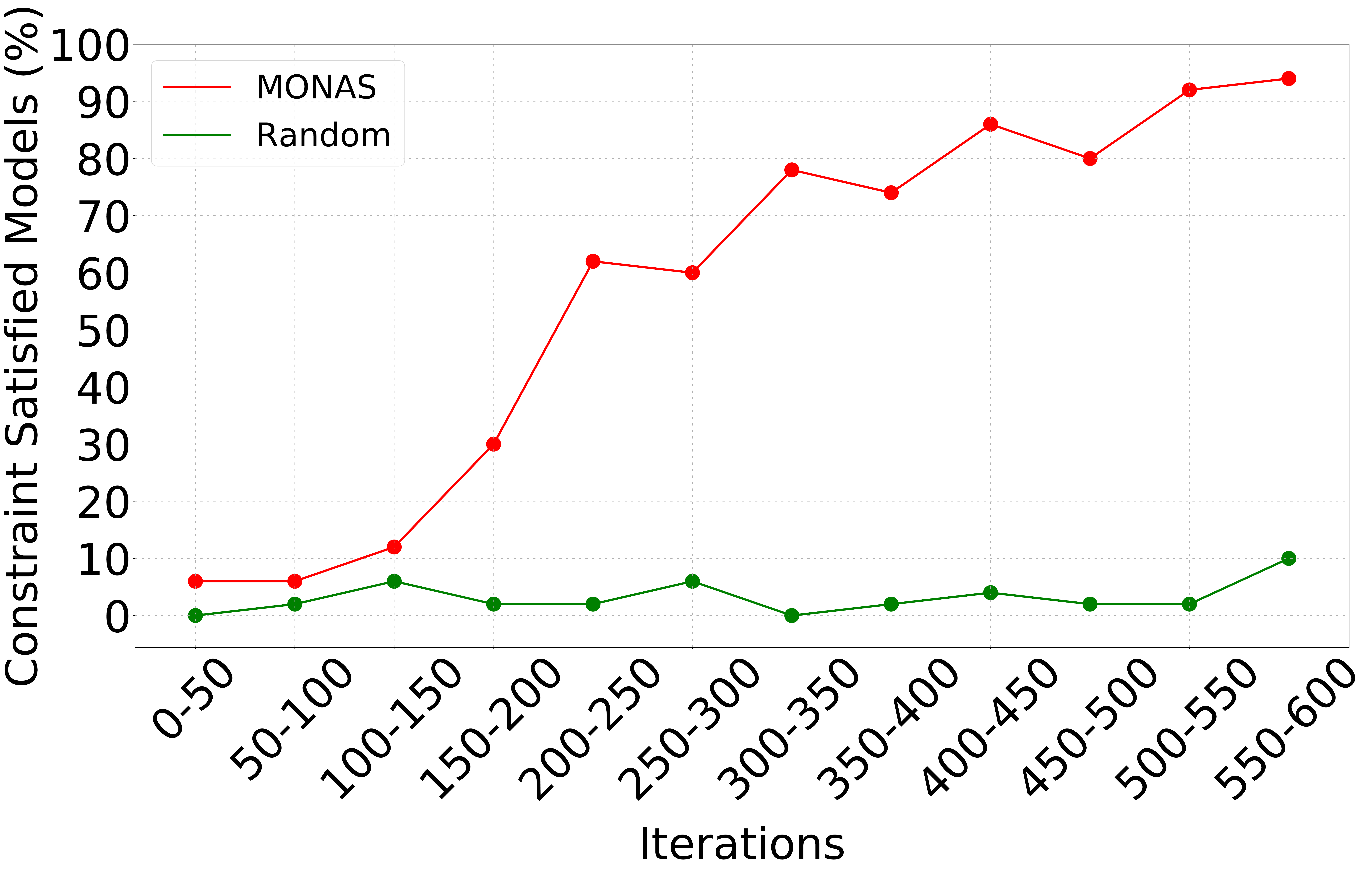}
\label{fig:70vsRan}}
\subfigure[Min Accuracy 0.85 vs Random Search]
{\includegraphics[width=0.45\textwidth]{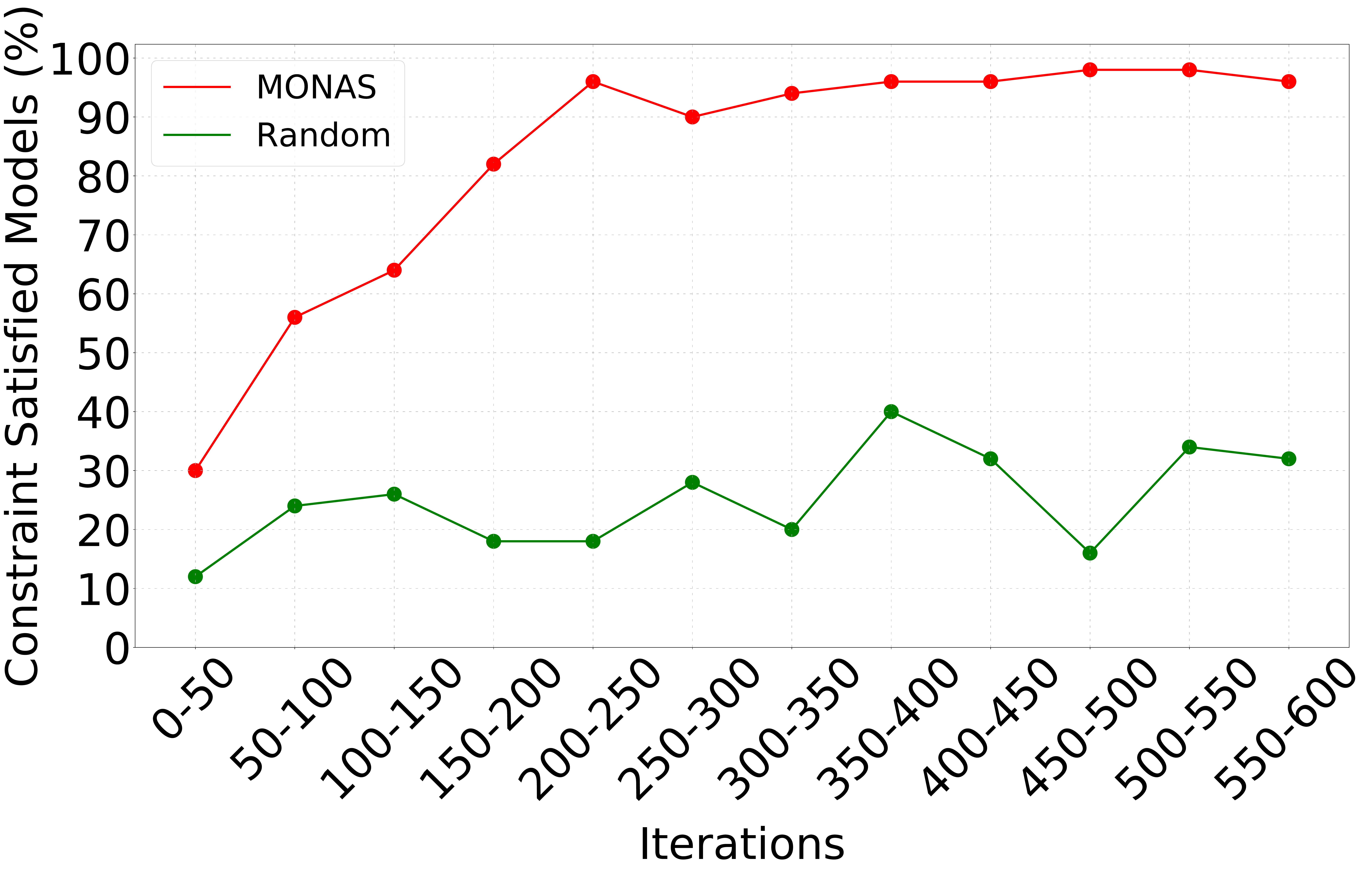}
\label{fig:0.85vsRan}}
\caption{ \small \monas efficiently guides the search toward child networks satisfying constraints. X axis represents the number of iterations that \rn has been trained, and Y axis represent the percentage of child networks satisfying constraints. In the beginning \monas has similar behaviors as random search; after 500 iterations, \monas almost always generates child networks satisfying constraints.}
\vspace{-5mm}
\label{fig:effchart}
\end{figure}

\begin{figure}[H]
\centering
\vspace{-5mm}
\subfigure[$\alpha=0.25$ (focus on energy efficiency)]
{\includegraphics[width=0.45\textwidth]{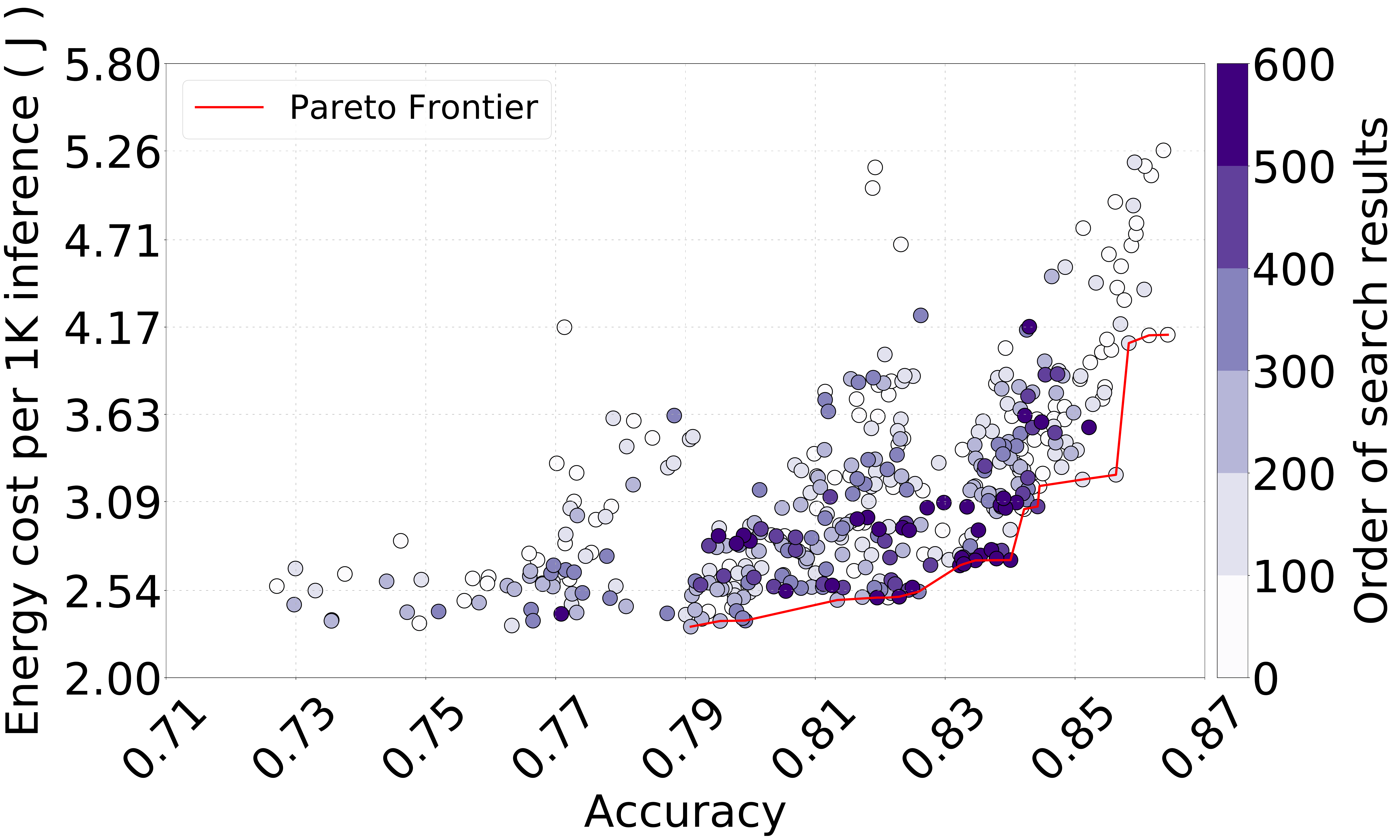}
\label{fig:mix0.25_alex_net}}
\subfigure[$\alpha=0.75$ (focus on accuracy)]
{\includegraphics[width=0.45\textwidth]{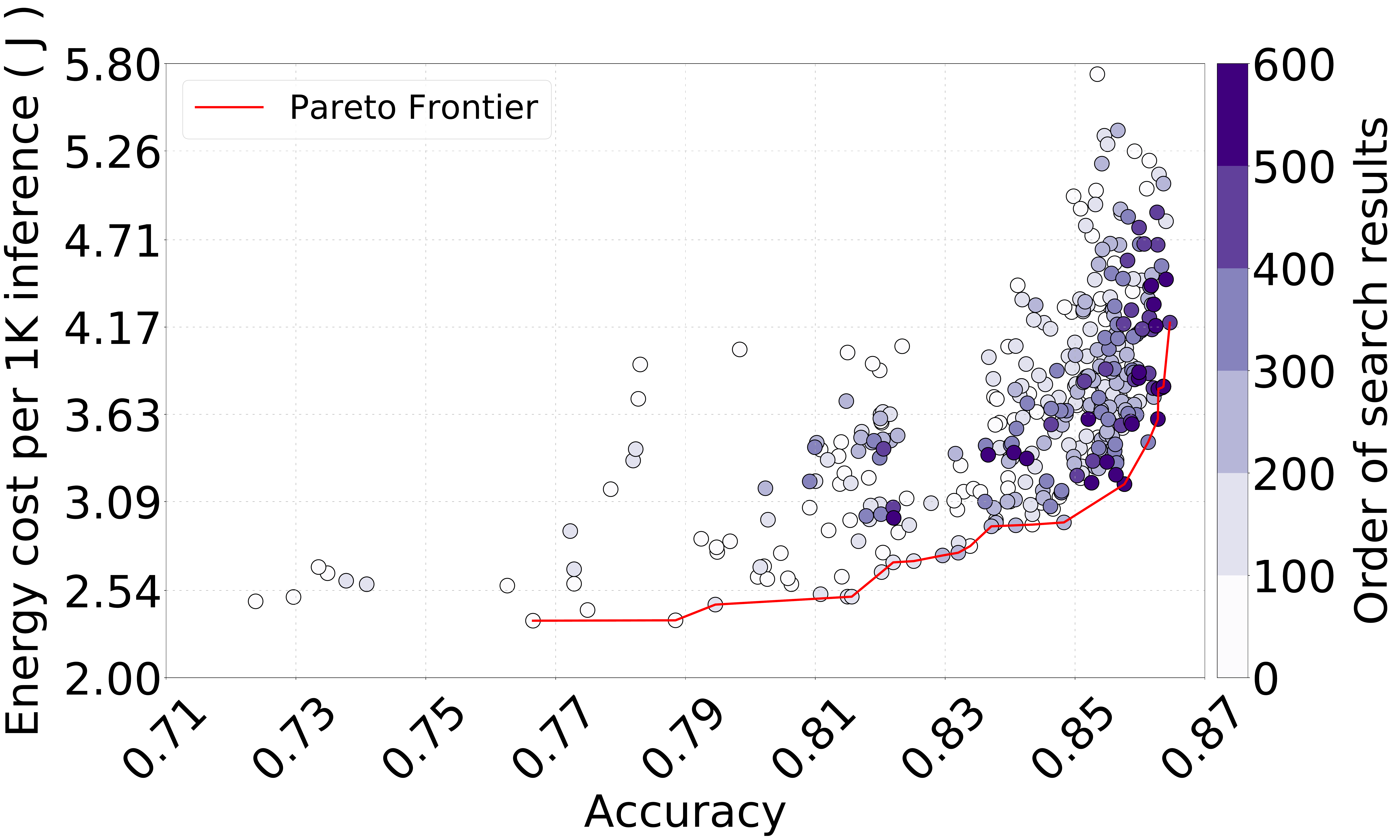}
\label{fig:mix0.75_alex_net}}
\caption{ \small Effects of different $\alpha$ values. X axis represents the accuracy and Y axis is the energy consumption. Each sample here is a generated child network. Notice that (a) when alpha is 0.25, most of the generated child networks have lower energy consumption, and (b) when alpha is 0.75, most of the generated network have higher accuracies. Red curves indicate the Pareto frontier.}
\vspace{-5mm}
\label{fig:alexnet_mix}
\end{figure}

\begin{figure}[H]
\centering
\vspace{-5mm}
\subfigure[Three Pareto frontiers achieved by: $\alpha = 0.25$, $\alpha = 0.75$, and random search]
{\includegraphics[width=0.45\textwidth]{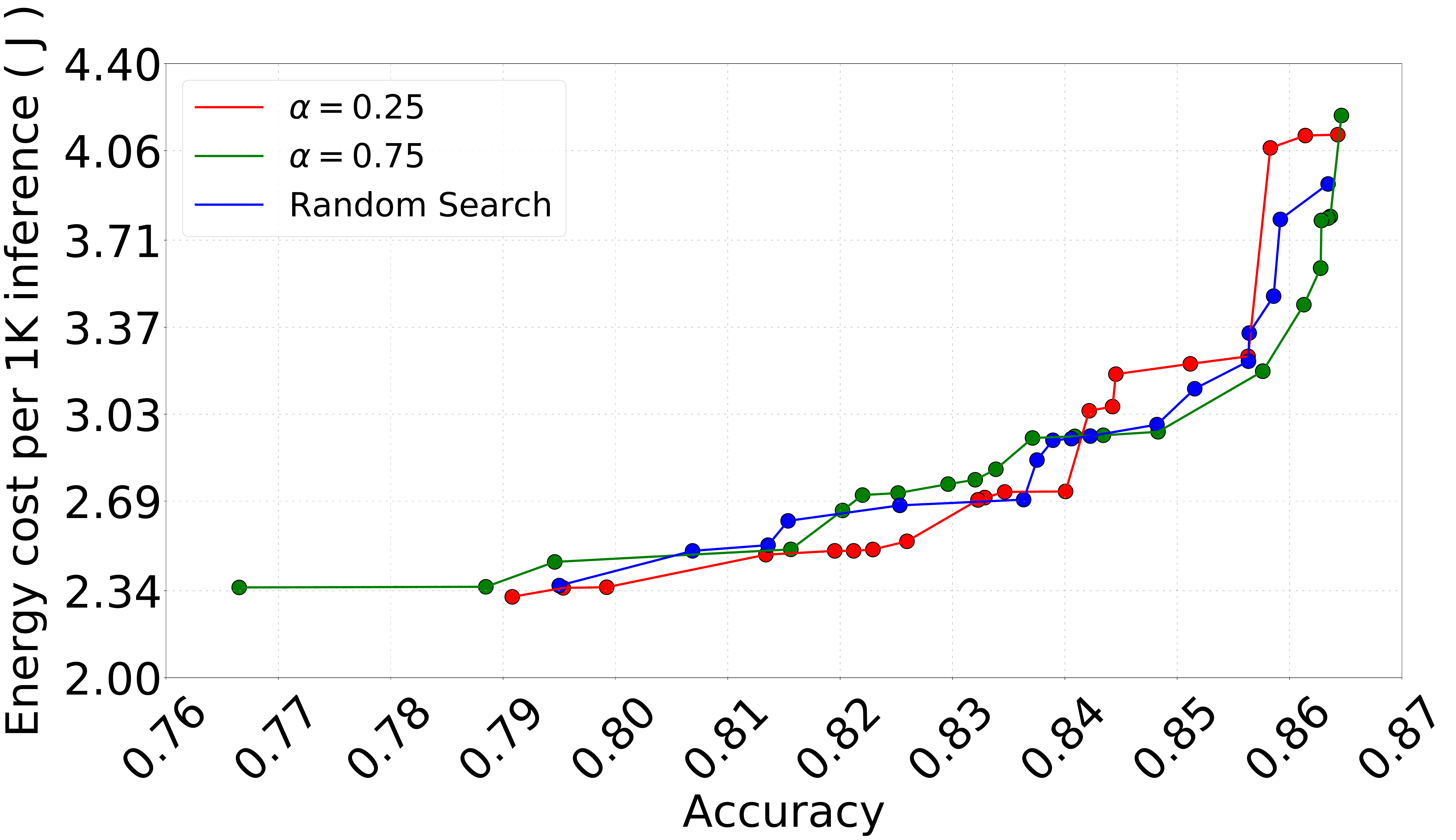}
\label{fig:paretos}}
\subfigure[Comparisons between \monas and CondenseNet.]
{\includegraphics[width=0.40\textwidth]{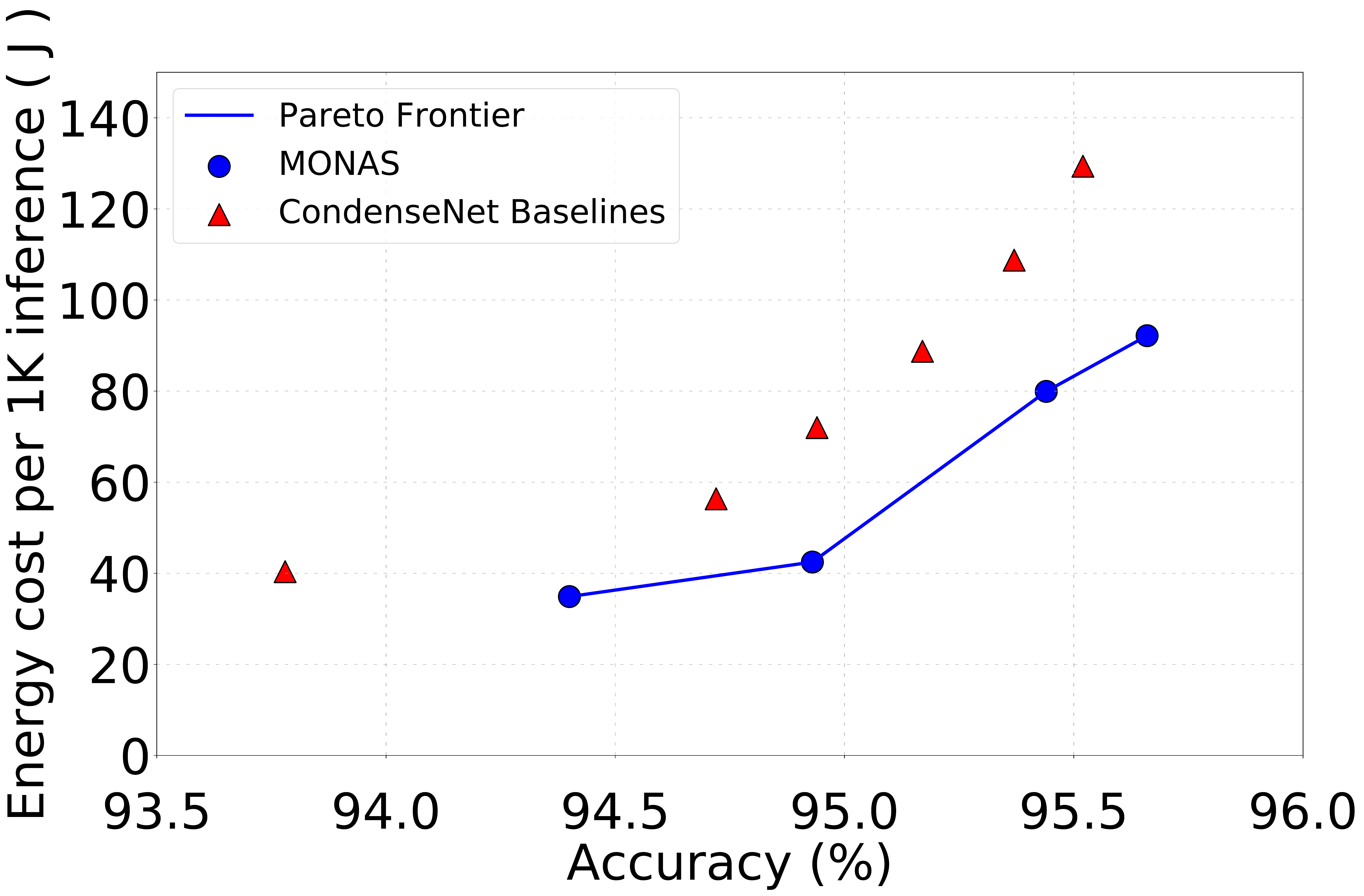}
\label{fig:p_front_condense}}
\caption{ \small Pareto Frontiers achieved by different NAS algorithms and models. X axis is the accuracy and Y axis represents the energy consumption. (a) The Pareto frontier achieved by random search is in-between the ones achieved by $\alpha = 0.25$ and $\alpha = 0.75$, showing that \monas is effective in searching for more energy-efficient or accurate models. (b) The Pareto frontier achieved by 
\monas outperforms CondenseNet, showing \monas is able to find a better model in terms of both accuracy and energy-efficiency.}
\vspace{-5mm}
\label{fig:paretos_analysis}
\end{figure}

Fig.~\ref{fig:effchart} shows the reward functions proposed by \monas efficiently guide NAS to focus on certain region of the search space for generating child networks that almost always satisfy constraints. Fig.~\ref{fig:alexnet_mix} shows the effects of applying different $\alpha$ values to the reward function in Eq~\eqref{eq:mix}. With different $\alpha$ values, \monas searches for different regions in the architectural space, leading to more energy-efficient networks in Fig.~\ref{fig:mix0.25_alex_net}, or more accurate networks in Fig.~\ref{fig:mix0.75_alex_net}. Fig.~\ref{fig:paretos_analysis} provides the analyses of Pareto frontiers achieved by \monas, random search, and CondenseNet. Experimental results confirm the effectiveness of \monas. Specifically, in Fig.~\ref{fig:p_front_condense}, the best model found by \monas is compared with the best one selected from~\citep{huang2017condensenet}. The Pareto Frontier in Fig.~\ref{fig:p_front_condense} demonstrates that the models found by \monas have both higher accuracy and lower energy compared to the model designed by domain experts. 

\pdfoutput=1
\section{EA-like Multi-Objective NAS}
\label{sec:dpp}

In this section, we introduce an EA-like approach for multi-objective NAS: \dpp~\cite{dong2018dpp}. \dpp searches for neural networks with a predefined number of cells to achieve Pareto-optimal performance over multiple objectives. Each cell contains multiple layers, and each layer has the type ``normalization (Norm)'' or ``convolutional (Conv).'' \dpp progressively adds layers following the order: Norm-Conv-Norm-Conv (repeat). \dpp's search space covers hand-crafted operations from MobileNet~\cite{howard2017mobilenets}, CondenseNet~\cite{huang2017condensenet}, ShuffleNet~\cite{zhang2017shufflenet} to take advantages of experts' knowledge on designing efficient CNNs. This improves the quality of each generated child network.

\subsection{Search Algorithm}
\label{sec:sec:searchalgo}

\begin{figure}[H]
\begin{center}
\includegraphics[width=0.75\linewidth]{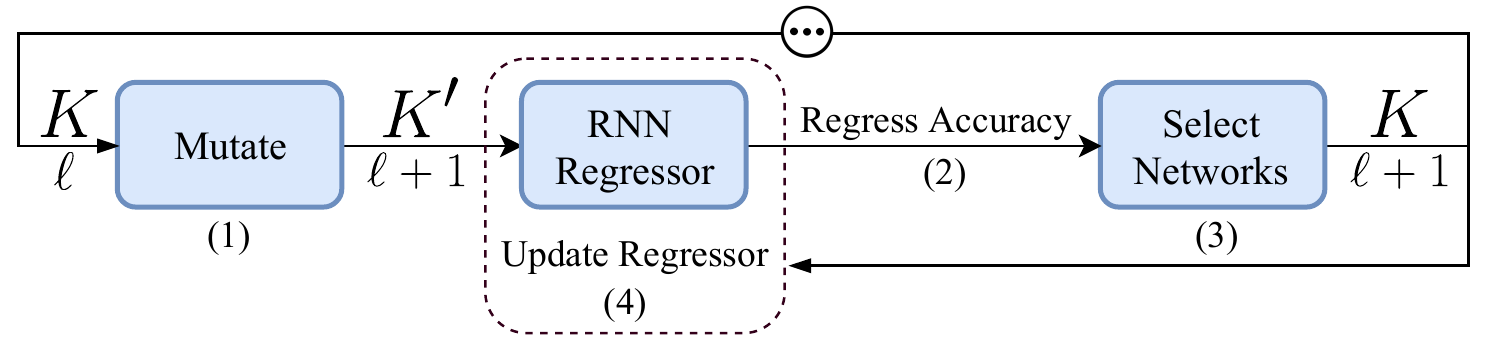}
\end{center}
\caption{\small Diagram of the search algorithm in \dpp (from Dong et al.~\cite{dong2018ppp}). The search process has four steps: (1) mutate to find child networks, (2) predict their accuracies, (3) select the networks achieving Pareto optimality to mutate for the next generation, and (4) update the regressor/predictor.}
\label{fig.diagram}
\end{figure}

Inspired by \cite{liu2017progressive,hutter2011sequential}, Dong et al.~\cite{dong2018ppp} adopt Sequential Model-Based Optimization (SMBO), an EA-like algorithm that contains four steps (see Fig.~\ref{fig.diagram} for diagram):
\begin{enumerate}
\item{\textit{\textbf{Mutate}}.} For each $\ell$-layers block, they enumerate all possible $\ell+1$-layers blocks. $K$ is the number of models to train, and ${K}'$ is the number of models after mutation.

\item{\textit{\textbf{Regress accuracy}}.} \dpp uses a Recurrent Neural Network (RNN) to predict a child network's accuracy given its architecture with zero training.
\item{\textit{\textbf{Select networks}}.} \dpp's main contribution is to use Pareto Optimality over multiple objectives to select $K$ networks, shown in Fig.~\ref{fig.diagram} step (3), rather than simply selecting the top $K$ accurate ones as in \cite{liu2017progressive}.
\item{\textit{\textbf{Update regressor}}.} Each of the selected $K$ networks are trained for $N$ epochs. The child networks (as inputs) and their corresponding evaluation accuracies (as outputs) are used to update the RNN regressor.
\end{enumerate}

\subsection{Experimental Results}
\label{sec:sec:dpp-results}

To begin with, Dong et al.~\cite{dong2018ppp} conduct the \dpp search with CIFAR-10 dataset with standard augmentation. After the search is done, the cell structure is used to form a larger model; then this model is trained and evaluated for ImageNet \cite{deng2009imagenet} classification. All the experimental results in this section are excerpted from the original \dpp paper (Dong et al.~\cite{dong2018ppp}).

Table.\ref{tb.hwsettings} provides the details of the devices considered by \dpp during the search process. Mainly, there are three types of devices considered: workstation (WS), embedded system (ES) and mobile phones (M). When searching models for WS and ES, four objectives are considered: error rate, number of parameters, FLOPs, and actual inference time on the devices. When search models for mobile phones, memory usage is also included as the $5^{th}$ objective.

\begin{table}[h]
\caption{\small \textbf{Hardware Specifications of Devices in \dpp.}}
\label{tb.hwsettings}
\centering
\begin{tabular}{|c|c|c|c|}
\hline
           & \textbf{Workstation (WS)} & \textbf{Embedded System (ES)} & \textbf{Mobile Phone (M)} \\ \hline
Instance   & Desktop PC           & NVIDIA Jetson TX1  & Xiaomi Redmi Note 4           \\ \hline           
CPU        & Intel i5-7600        & ARM Cortex57      & ARM Cortex53               \\ \hline
Cores      & 4                    & 4                   & 8                            \\ \hline
GHz        & 3.5                  & 1.9                 & 2.0                          \\ \hline
CUDA       & Titan X (Pascal)     & Maxwell 256         & -                            \\ \hline
Memory     & 64 GB / 12 GB        & 4 GB                & 3 GB                         \\ \hline
Objectives & 4                    & 4                   & 5                            \\ \hline
\end{tabular}
\end{table}

\begin{table*}[h]
\caption{\small \textbf{Cifar10 classification results.}
Missing values are the metrics not reported by the original papers. The standard deviations reported in the bottom row for DPP-Net-Panacea are calculated across 10 runs. Numbers in bold font highlight the smallest value of each column.}
\label{tb.cifar}
\centering
\resizebox{1\columnwidth}{!}{%
\begin{tabular}{l|lll|llll}
\hline
                                   & \multicolumn{3}{c|}{\textit{Device-agnostic metrics}} & \multicolumn{4}{c}{\textit{Device-aware metrics}}    \\ \hline
\textbf{Model from previous works} & Error rate     & Params      & FLOPs       & Time-WS & Time-ES & Time-M & Mem-M \\ \hline
Real et al. \cite{real2017large}   & 5.4              & 5.4M        & -           & -       & -       & -           & -          \\
NASNet-B \cite{zoph2017learning}   & 3.73             & 2.6M        & -           & -       & -       & -           & -          \\
PNASNet-1 \cite{liu2017progressive}& 4.01             & 1.6M        & -           & -       & -       & -           & -          \\ \hline
DenseNet-BC (k=12) \cite{huang2017densely} & 4.51     & 0.80M       & -           & -       & -       & 0.273       & 79MB       \\
CondenseNet-86 \cite{huang2017condensenet} & 5.0      & 0.52M       & 65.8M       & 0.009   & 0.090   & 0.149       & 113MB      \\ \hline
                                   & \multicolumn{3}{c|}{\textit{Device-agnostic metrics}} & \multicolumn{4}{c}{\textit{Device-aware metrics}}    \\ \hline
\textbf{Model from DPP-Net}        & Error rate       & Params      & FLOPs       & Time-WS & Time-ES & Time-M & Mem-M \\ \hline
DPP-Net-PNAS     & \textbf{4.36}    & 11.39M      & 1364M       & 0.013   & 0.062   & 0.912       & 213MB      \\ \hline
DPP-Net-WS     & 4.78             & 1.00M       & 137M        & \textbf{0.006} & 0.075   & 0.210       & 129MB      \\
DPP-Net-ES     & 4.93             & 2.04M       & 270M        & 0.007   & \textbf{0.044} & 0.381       & 100MB      \\
DPP-Net-M & 5.84             & \textbf{0.45M} & \textbf{59.27M} & 0.008   & 0.065   & \textbf{0.145}  & \textbf{58MB}       \\ \hline
DPP-Net-Panacea  & 4.62 $\pm$ 0.23      & 0.52M       & 63.5M       & 0.009 $\pm$ 7.4e-5  & 0.082 $\pm$ 0.011   & 0.149 $\pm$ 0.017       & 104MB     
\end{tabular}
}
\end{table*}

\indent\textit{\textbf{Results on CIFAR-10.}}
Table~\ref{tb.cifar} provides the performance comparisons among \dpp, previous NAS literatures~\cite{zoph2017learning,real2017large,liu2017progressive} and the state-of-the-art handcrafted mobile CNN models: DenseNet-BC and CondenseNet~\cite{huang2017condensenet,huang2017densely}. Among all networks, DPP-Net-Panacea (bottom row) has a small error rate and performs relatively well on every objective. We also include the results of the neural network found by DPP-Net with PNAS \cite{liu2017progressive} criterion: using classification accuracy as the only objective. DPP-Net-PNAS has a large number of parameters and the corresponding inference time is slow; yet, it achieves a smaller error rate. This shows \dpp provides a flexible and effective framework to search for device-aware, high-performance neural networks. 

\begin{table*}[!h]
\caption{\small \textbf{ImageNet classification results.}
Time-ES represents the inference time on the embedded system; Time-M and Mem-M are the inference time and memory usage, respectively, when the model is deployed on a mobile phone.}
\label{tb.imagenet}
\centering
\begin{tabular}{l|lllllll}
\textbf{Model} &\textbf{Top-1} &\textbf{Top-5} &\textbf{Params} &\textbf{FLOPs}  &\textbf{Time-ES} &\textbf{Time-M} &\textbf{Mem-M}  \\ \hline
Densenet-121 \cite{huang2017densely}           & 25.02	 & 7.71 & -	     & -	   & 0.084  & 1.611  & 466MB   \\
Densenet-169 \cite{huang2017densely}           & 23.80   & 6.85 & -	     & -	   & 0.142  & 1.944  & 489MB   \\
Densenet-201 \cite{huang2017densely}           & 22.58	 & 6.34 & -	     & -	   & 0.168  & 2.435  & 528MB   \\ \hline
ShuffleNet 1x (g=8) \cite{zhang2017shufflenet} & 32.4    & -    & 5.4M   & 140M    & 0.051  & 0.458  & 243MB   \\
MobileNetV2 \cite{sandler2018inverted}         & 28.3    & -    & 1.6M   & -       & 0.032  & 0.777  & 270MB   \\ 
Condensenet-74 (G=4)\cite{huang2017condensenet}& 26.2	 & 8.30 & 4.8M   & 529M    & 0.072  & 0.694  & 238MB   \\ \hline
NASNet (Mobile) \cite{zoph2017learning}        & 26.0	 & 8.4	& 5.3M   & 564M    & 0.244  & -      & -       \\ \hline
DPP-Net-PNAS                                   & 24.16   & 7.13 & 77.16M & 9276M   & 0.218  & 5.421  & 708MB   \\ 
DPP-Net-Panacea       & 25.98   & 8.21 & 4.8M   & 523M    & 0.069  & 0.676  & 238MB   \\ 
\end{tabular}
\end{table*}

\indent\textit{\textbf{Results on ImageNet}}. Table~\ref{tb.imagenet} provides the comparisons among \dpp and other models on the ImageNet classification task. Notice that DPP-Net-Panacea outperforms the state-of-the-art, hand-crafted architecture Condensenet-74 in almost every aspect. Moreover, DPP-Net-Panacea also outperforms NASNet (Mobile), a single-objective, state-of-the-art NAS approach~\cite{zoph2017learning} in every objective. These results demonstrate that high model accuracy and device-related objectives (e.g., latency) can be achieved/optimized at the same time, without compromising one over the others.

\section{Discussions}
\label{sec:discussions}

In this paper, we survey recent literatures on NAS and summarize them into four categories: (a) RL, (b) EA, (c) search acceleration and (d) multi-objective. Also, currently there are two main streams of designing the search space for NAS: covering the entire network (macro) or only one cell (micro). 

We also deep dive into two recent advancements of multi-objective NAS: \monas and \dpp. \monas adapts rewards to application-specific constraints and effectively guide the search process to find the models of interest, such as achieving higher accuracy and at the same time lower energy consumption.

To the best of our knowledge, \dpp is the first device-aware NAS outperforming state-of-the-art handcrafted mobile CNNs. Experimental results on CIFAR-10 demonstrate the effectiveness of Pareto-optimal networks found by \dpp, for three different devices: (a) a workstation with NVIDIA Titan X GPU, (b) NVIDIA Jetson TX1 embedded system, and (c) mobile phone with ARM Cortex-A53. Compared to CondenseNet and NASNet (Mobile), \dpp achieves better performances: higher accuracy \& shorter inference time on these various devices. 

Most of the successes achieved by NAS (both single and multiple objectives) are in convolutional neural networks and image-related domains. Therefore, we believe exploring NAS into other domains is a straight-line and important future work. Furthermore, all previous works on multi-objective NAS adapt existed search space and acceleration methods. In other words, no search space or acceleration methods are proposed specifically for multi-objective NAS. Given the importance and high complexity of multi-objective NAS, we believe designing search space or acceleration method will be a critical and challenging future direction in this field.

\bibliographystyle{unsrt}

\end{document}